\documentclass[letterpaper, 10 pt, conference]{ieeeconf}  
\usepackage{stdlib}

\IEEEoverridecommandlockouts                              

\title{\LARGE \bf
\approach{}: A Framework for Benchmarking Social Robot Navigation
}

\author{Jarrett Holtz$^{1}$ and Joydeep Biswas$^{1}$
\thanks{$^{1}$Computer Science, University of Texas, Austin, TX, 78712, USA
        {\tt\small jaholtz@utexas.edu}, {\tt\small joydeepb@cs.utexas.edu}}%
}

\begin{document}

\maketitle
\thispagestyle{empty}
\pagestyle{empty}

\begin{abstract}
  Robots moving safely and in a socially compliant manner in dynamic human
  environments is an essential benchmark for long-term robot autonomy.
  However, it is not feasible to learn and benchmark social navigation behaviors
  entirely in the real world, as learning is data-intensive, and it is challenging
  to make safety guarantees during training.
  Therefore,
  simulation-based benchmarks that provide abstractions for social navigation
  are required.
  A framework for these benchmarks would need to support a wide variety of
  learning approaches, be extensible to the broad range of social navigation
  scenarios, and abstract away the perception problem to focus on social navigation
  explicitly.
  While there have been many proposed solutions, including high fidelity
  3D simulators and grid world approximations, no existing solution
  satisfies all of the aforementioned properties for learning and evaluating social
  navigation behaviors.
  In this work,
  we propose \approach{}, a lightweight 2D simulation environment
  for robot social navigation designed with extensibility in mind, and a
  benchmark
  scenario built on \approach{}.
  Further, we present benchmark results that compare and contrast
  human-engineered and model-based learning approaches to a suite of
  off-the-shelf
  Learning from Demonstration (LfD) and Reinforcement Learning (RL) approaches applied
  to social robot navigation. These results demonstrate the data efficiency,
  task performance, social compliance, and environment transfer capabilities
  for each of the policies evaluated to provide a solid grounding
  for future social navigation research.
\end{abstract}




\section{Introduction}
\seclabel{intro}
Deploying robot navigation safely alongside people such that they move
in a social manner is one of the ultimate goals of robotics.
However, training
and evaluating in real-world human environments present both safety concerns
and scalability challenges for many learning algorithms, and this difficulty compounds
the difficulty of developing robust approaches to robot social navigation.
As such,
an extensible benchmark for social navigation is a critical
step along the path to deploying socially compliant robots.

In order to accommodate the rapidly expanding pool of promising machine learning
techniques a framework for these benchmarks should provide the infrastructure
for integrating approaches.
In addition, the ability to adjust
the reward, observation space, and properties of the agent and environment
are important for capturing social navigation scenarios and
enabling varied learning approaches.
Finally, such a framework should provide a representation with
reasonable simulation fidelity while also abstracting away the perception
task to focus on social navigation.
Many different solutions for simulating robots and humans in dynamic
environments
have been proposed for social navigation research. However,
to the best of our knowledge no existing solution satisfies all of the desired
properties for a benchmark framework.

\begin{figure}
  \centering
  \includegraphics[width=\columnwidth]{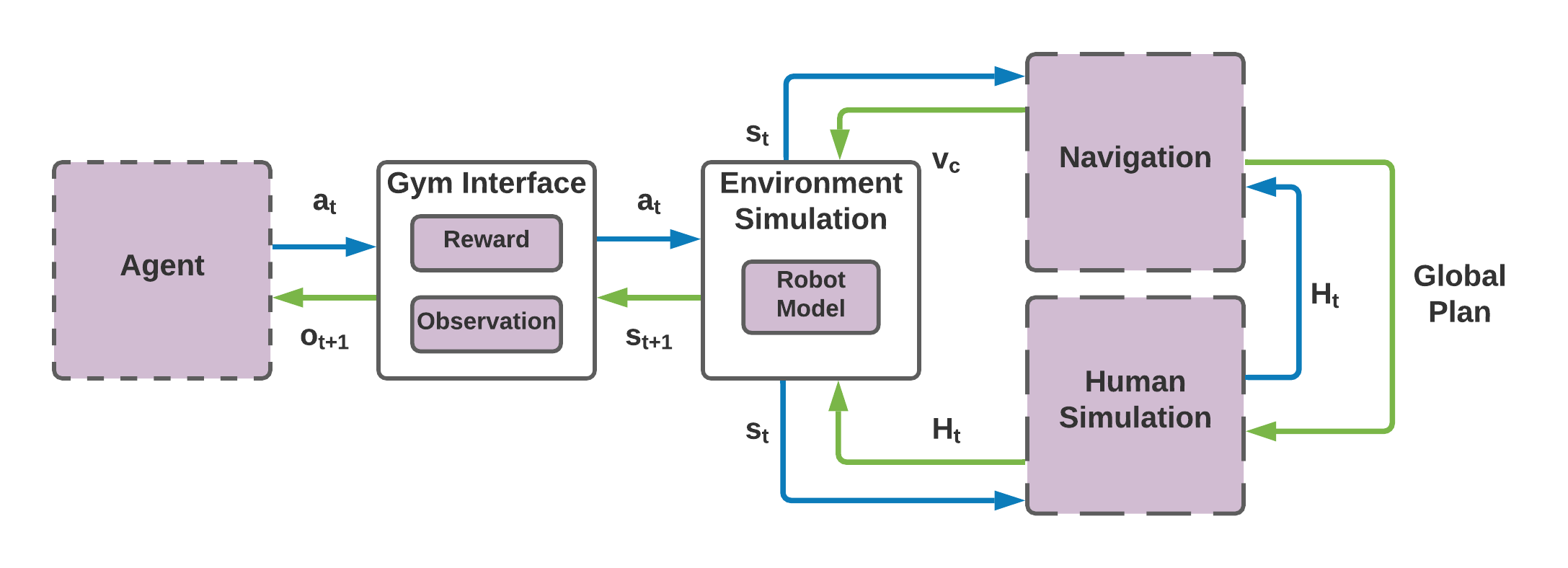}
  \vspace{-1em}
  \caption{Diagram of interaction between modules in \approach{}. Dashed boxes
  represent interchangeable modules, while purple boxes contain configureable
  parameters.
  Blue arrows represent requests between modules, and green arrows represent responses.}
  \figlabel{systemDiagram}
  \vspace{-2em}
\end{figure}

Our proposed solution is \approach{}, a 2D simulation environment built on the
Robot Operating System (ROS) to provide a lightweight
and configurable option for training and evaluating social navigation
behaviors: https://github.com/ut-amrl/social\_gym
. 2D simulation is chosen to provide an
option that abstracts away the perception problem to focus on interactions
between navigation and people in dynamic human environments.
To streamline
integration in common reinforcement learning workflows, \approach{} implements
the OpenAI gym interface for training and evaluation. \approach{} is modular
such that modules can be exchanged for others to increase the range of
possible benchmarks, an overview of module interactions and configurability
is shown in \figref{systemDiagram}, and a visualization of the simulation
environment in \figref{introFigure}.
The included benchmark
in \approach{} focuses on the action selection problem presented in work on
Multipolicy Decision Making \cite{mpdm2} and program synthesis for social navigation
\cite{holtz2021idips}, where
social navigation is modelled as an action selection problem, and
the optimal action is selected from a set of discrete sub-policies.
Using this included benchmark, we compare and contrast a suite of
engineered approaches,
model-based symbolic learning, Learning from Demonstration (LfD), and
Reinforcement Learning (RL); to provide a baseline for future comparisons;
and highlight promising future directions for research. Our results demonstrate
the data efficiency, scenario performance, social compliance, and generalizability
of different policies, and our analysis highlights the promising strengths
and weaknesses of different techniques.
In addition to providing
the benchmark for multipolicy decision-making in social navigation,
\approach{} is designed to be configurable
and extensible in the hopes of providing a stable base for a variety of
social navigation benchmarks.

The contributions of this work are as follows:
\begin{itemize}
    \item a 2D simulation environment for robot social navigation,
    \item a complete toolkit that includes robot simulation, human simulation,
      navigation stack, and simple baseline behaviors,
    \item OpenAI gym integration to provide a standard interface for many learning
      approaches,
    \item a benchmark based on multipolicy decision making for
      robot social navigation, and
    \item evaluation results and analysis for several reinforcement learning
      and learning from demonstration approaches for use as baselines for future research.
\end{itemize}

\section{Related work}
\seclabel{related}
Model-based approaches with application-specific engineered behaviors have
historically been applied to social navigation
\cite{mpdm2, socialForce, socialForce2, proxemics, socialMaps}.
More recently, approaches
have sought to leverage machine learning techniques to
learn general social navigation behaviors
\cite{how2017, hoof2020, alahi2019, fu2018, burgard2018, scand}.
Prior work has explored the use of program synthesis and learning from demonstration
to learn social navigation behaviors as symbolic programs~\cite{holtz2018interactive, holtz2020ldips, holtz2021idips}.

Existing simulators and benchmarks are either at the level of high-fidelity
3D simulation that focuses on the raw perception problem of humans in the
environment or are simplified grid world models of navigation. In contrast,
\approach{} provides sensor reading at the level of laser scans and human
detections.


Many general-purpose high-fidelity 3D simulators have been developed
for robotics. Of particular
note is the Gazebo simulator~\cite{gazebo}, a powerful and extensible simulator commonly
used with ROS. Other high-fidelity simulators have been developed
using gaming physics and graphics engines such as Unity~\cite{unity} or Unreal~\cite{unrealengine}
with autonomous driving in mind, some of which can simulate other agents in the
environment, such as AirSim~\cite{airsim} and CARLA~\cite{carla}.
In addition to these more general-purpose simulators, some solutions have
been proposed specifically for the social navigation problem. The
Social Environment for Autonomous Navigation (SEAN)~\cite{sean} uses a similar
Unity-based approach to more general simulators while providing social
navigation-specific metrics and scenarios. While SEAN does not provide any
particular benchmark, SocNavBench~\cite{socnavbench} is designed as a social benchmark
that generates photo-realistic sensory input directly from pre-recorded
real-world datasets to strike a balance between simulation and recorded datasets.
In addition to simulation-based benchmarks, a small number of purely
dataset-based benchmarks have been proposed, including SocNav1~\cite{socnav1}
and Social Robot Indoor Navigation (SRIN)~\cite{srin}.
While datasets are a critical benchmark tool, they often require an initial sensor
processing step and are not easily extensible.
Alternatively, simplified grid-world type environments, such as
MiniGrid~\cite{minigrid}, can also be used to approximate social
navigation by modelling humans with dynamic obstacles, although this is
not commonly used for benchmarking.

\section{Social Navigation in Robotics}
\seclabel{background}
We frame social navigation as a discounted-reward partially observable
Markov Decision Process (POMDP)
$M = \langle S, A, T, R, \Omega, O, \gamma \rangle$ consisting of
the state space $S$ where $s \in S \mapsto \langle p_r, v_r, p_g, H \rangle$ and
$p_r$ is the robot pose, $v_r$ is the robot velocity, $p_g$ is the goal pose,
and $h_i \in H \mapsto \langle p_{hi} v_{hi} \rangle$ is a list of the
human poses and velocities;
actions $A$ represented either as discrete motion primitives~\cite{mpdm2} or
continuous local planning actions~\cite{costFunctions}; the world transition
function
\begin{equation}
  T(s, a, s') = P(s_{t + 1} = s' \given s_t = s, a_t = a)
\end{equation}
for the probabilistic transition to states $s'$ when taking action $a$ at
previous state $s$; the reward function  $R : S \times A \times S \mapsto \real$
; the set of observations $o \in \Omega$; a set of conditional
observation probabilities $O$; and discount factor $\gamma$.
The solution to this POMDP is represented as a
policy $\pi : S \times A \mapsto A$ that decides what actions to take
based on the previous state-action pair. The optimal social
navigation policy $\pi^*$ maximizes the expectation over the cumulative
discounted rewards:
\begin{align}
  \pi^* &= \arg_{\pi}\max J_\pi\nonumber,\\
  J_\pi &= E \left[ \sum_{t=0}^{t=\infty} \gamma^t R(s_t, \pi(s_t, a_t),
  s_{t+1}) \right]
\end{align}
We next describe how \approach{} provides modules to simulate the components
of the POMDP and interfaces with different approaches to learn
$\pi^*$.


\section{\approach{} System architecture}
\approach{} is built from four modules that simulate different components of the POMDP
for social robot navigation and coordinate together to simulate the full
POMDP. These modules are the \textbf{Gym} module, the \textbf{Environment}
module, the \textbf{Human} module, and the \textbf{Navigation} module.

The Gym module is the top-level interface between the agent and the simulator
that handles simulation of the agent's action selection
policy to select an action $a_t \in A$ at each timestep $t$ based on the
current observation
$o_t \in \Omega$. The Gym module then steps the simulation by passing $a_t$
to the Environment module, which returns a new state $s_{t+1} \in S$
from which the Gym module calculates the reward
$r_t$ and derives a new observation $o_{t+1} \in \Omega$.
The Environment module handles the representation of the state $s_t \in S$ and coordinates
with the Human and Navigation modules to simulate the transition function $T$
given $a_t$ at each timestep.
The Human module takes in $s_t$ and $a_t$ and controls the components of
$T$ governed by the humans' response to the environment by sending updated
human positions and velocities to the Environment module. In turn,
the Navigation module controls the components of $T$ related to robot execution
of $a_t$ by providing a baseline navigation behavior for evaluation
and global and local planners for use in discrete action spaces.

While our current implementation provides a
single complete benchmark scenario and extensions to enable various
others, it is essential to note that each module can be freely replaced with
any alternative that implements a prerequisite ROS service. In this way,
ROS services make our approach accessible to a wide variety
of existing approaches written in ROS.
The components of \approach{} and their interactions are shown in
\figref{systemDiagram}, and in the following sections, we will describe
the abstractions of each module and high-level technical details,
while in \secref{benchmark}, we will describe the concrete instantiation used
for our included benchmark.

\begin{figure}
  \centering
  \includegraphics[width=\columnwidth]{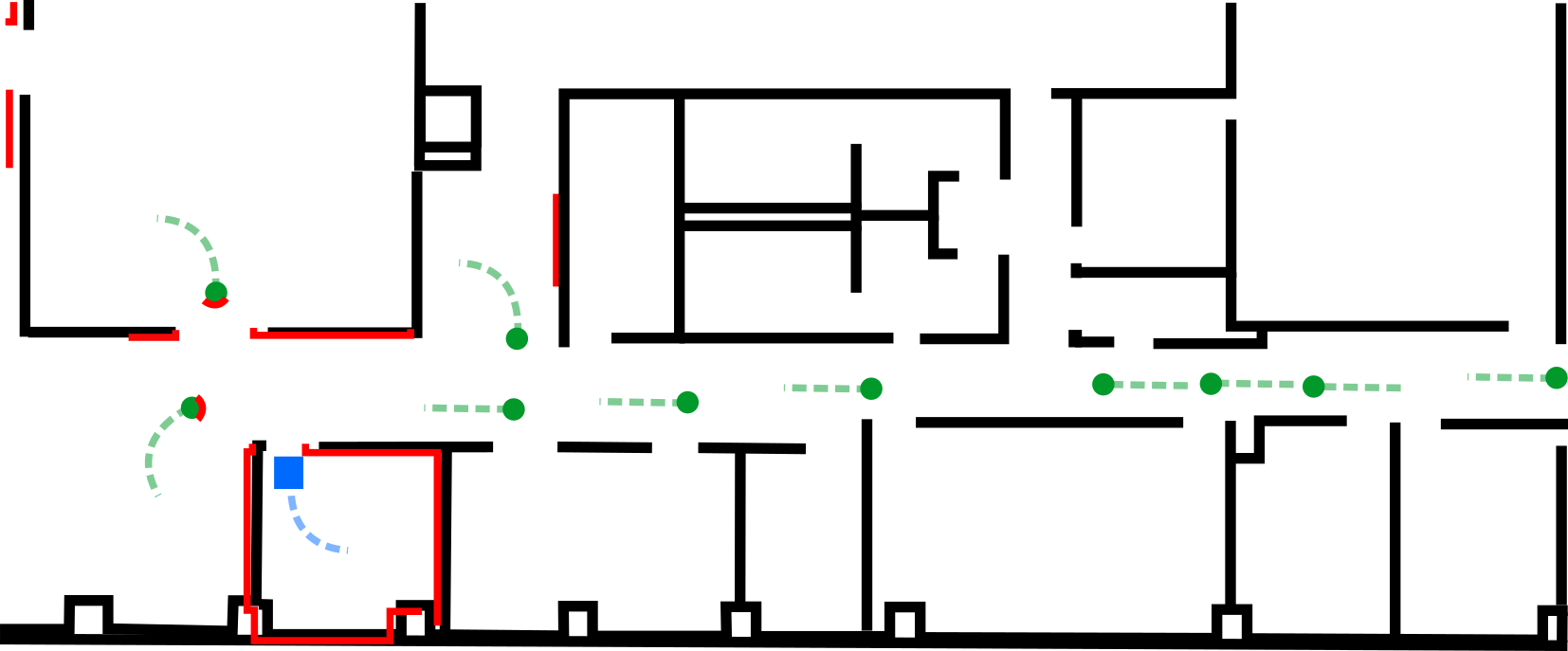}
  \vspace{-1em}
    \caption{Visualization of one timestep of \approach{} with humans shown as green
    circles, the walls as black lines, the laserscan as redlines, the robot as a blue box, and recent
    trajectories as dotted lines.}
  \figlabel{introFigure}
  \vspace{-2em}
\end{figure}

\subsection{Gym Integration}
\seclabel{gym}
The Gym model builds on the OpenAI Gym framework to interface
between our ROS-based simulation framework and the learning agent. This module
has four responsibilities: coordinating the generation and resetting of scenarios,
receiving actions $a \in A$ from the
agent that are used to coordinate state transition of the environment, receiving
states $s \in S$ from the environment and converting them into observations
$o \in \Omega$, and evaluating performance by calculating metrics and the
value of the reward function $R \given s, a$.

\paragraph{Generating scenarios}
A scenario is described as a tuple $\langle m, t, p_s, p_g, H \rangle$, where
$m$ is the map describing the static obstacles in the scene,
$t$ is the timestep,
$p_s$ is the start
pose of the robot, $p_g$ is the goal pose of the robot,
and $H$ is a list of pairs $\{\langle p_{h}^{i}, p_{g}^{i}\rangle \}$ that describes the initial pose $p_{h}^{i}$
and the goal pose $p_{g}^{i}$ for each human, $i$, in the scene. \approach{} randomly generates
new scenarios based on several parameters that describe the range of possible
initial conditions and goals.
These parameters are $m$, the maximum and minimum number of humans
$h_{max}, h_{min}$, the number of iterations
for a given scene $N_r$, and the maximum total iterations $N_m$.
In addition, scenario generation requires a list of
poses $G$ that describes the list of legal start
and end locations in the map.
Given this configuration, a new scenario is generated as follows:
\begin{paraenum}
  \item $m$ and $G$ are configured by the user,
  \item a random $p_s$ and $p_g$ are selected from $H$,
  \item a random $n$ is chosen such that
        $n_{min} < n < n_{max}$,
      \item $n$ humans are generated by selecting a random $p_{h}^{i}$ and $p_{g}^{i}$
    from $G$ such that they do not overlap,
  \item and finally, the environment is initialized and the initial state
    is sent to the agent.
\end{paraenum}
Each random scenario is run until it completes in either success or failure
$N_r$ times, and then a new scenario is generated up to the maximum number
of iterations $N_m$. We define the success state to be when
$|p_r - g| <  \epsilon_s$, and the failure state to be when $t > t_f \land
|p_r - g| >  \epsilon_s$.

\paragraph{Updating the environment and observation}
The Gym module controls the update loop of the simulation as follows:
\begin{paraenum}
  \item the Gym module receives an action $a_t$ from the agent,
  \item $a_t$ is sent as a request to the environment,
  \item the Environment module executes $T$ given $a_t$,
  \item the Gym module receives a response state $s_{t+1}$,
  \item an observation function $f(s \in S) \mapsto o \in \Omega$ is employed to map
      $s_{t+1}$ to an observation $o_{t+1}$ where $o_{t+1}$ is an observation vector
    containing both discrete and continuous variables,
  \item the reward is calculated using $R(s_{t+1}) \mapsto r_t$, and finally
  \item $r_t$ and $o_{t+1}$ are passed to the agent for the next iteration.
\end{paraenum}
For use with, and in addition to calculating reward, the Gym module can optionally produce
several metric values for use in performance evaluation.
The provided Gym module supports the following metrics:
\begin{paraenum}
  \item time to goal: $t$ when the agent reaches $p_g$;
  \item distance from goal: $|p_{t+n} - g|$;
  \item distance traveled: $\sum_{t+1}^{t+n} |p_{t} - p_{t-1}|$;
  \item force: approximates maximum force exerted between the robot and humans
    $\argmax_i e^{|p_r - p_{h}^{i}|}$;
  \item blame: takes into account velocity at times of close encounters;
    penalizing robot velocities towards humans. Let $p_{r}^{i}$ be the closest point
    on the line segment from $p_{r}$ to $p_{r} + v_{r}$ to
    $p_{h}^{i}$, then blame is calculated as: $\argmax_i \Phi(|p_r - p_{h}^{i}|)$;
  \item human collision count, and
  \item static obstacle collision count.
\end{paraenum}


\subsection{Environment simulation}
The Environment module is responsible for representing
the state $s_t \in S$ and updating $s_t$ based on robot actions
$a_t$ and the state transition function $T$.

The environment updates the robot state according to the robot motion model
and $a_t$.
For discrete actions, $a_t$ must first be converted to continuous velocities
by the Navigation module.
In addition to updating the robot pose, the environment also needs to update
all humans in the scene. However, since $a_t$ does
not provide velocities for the humans, we need to describe how
humans will move in response to the environment. Here the
Human module~\secref{human} utilizes the updated state
from the environment to calculate new positions and velocities for
each human in the scene.


\subsection{Human simulation}
\seclabel{human}
The Human module is responsible for simulating the components of $T$ concerned
with how humans behave. This requires updating $H$ based on
some model of humans moving through the environment towards a goal.

Each human $h^i$ in the scene moves back and forth between a starting position
$h^{i}_{p}$, and a goal pose $h^{i}_{g}$, by planning a global path of intermediary
nodes between the two that avoids static obstacles and employs a separate local
planner to handle dynamic obstacles.
For global planning, we utilize the
Navigation module as described in \secref{navigation}.
The local planner \approach{} employs is PedsimROS~\cite{pedsim_ros}, a
ROS module that models human behaviors according to the
social force model~\cite{socialForce,socialForce2}.
In brief, the Social Force Model represents each agent in the environment
as exerting repelling force on each agent,
while goals exert attractive forces.
We refer the reader to the original work~\cite{socialForce}
and the PedsimROS implementation~\cite{pedsim_ros}.


\subsection{Navigation}
\seclabel{navigation}
The Navigation module serves three purposes. For general use, the
navigation module provides a global planning interface for use by the human
module and implementations of baseline deterministic behaviors that do not
require learning that can be used as comparisons.
For our included benchmark scenario,
the navigation module further provides implementations of \textit{actions}
that make up the discrete sub-policies of our action space.

The global planning component of Navigation is responsible for planning an
obstacle-free path from a start location to a goal location. Global planning
receives a request of the form $\langle m, p_s, p_f \rangle$, where
$p_s$ is a start position in the map $m$ and $p_f$ is the target
position and returns a
response of the form $\langle \{p_s, p_j, ... p_{j+n}, p_f \}\rangle$,
where $p_{j}$ represents an intermediary local goal along an obstacle-free
path from $p_s$ to $p_f$.

The local planner is responsible for avoiding obstacles immediately visible to
the robot and calculating control commands to take the robot to the
next waypoint on the global plan. The local planner is only called as part
of the complete navigation solution that receives a request of the form
$\langle m, p_s, p_f \rangle$,
and returns a response of the form $\langle v_c \rangle$, representing
command velocity to be executed by the Environment.
For the local planner, we implement
a trajectory rollout planner~\cite{trajectoryRollout} that
interfaces with the global planner to get the next local goal, and plans
obstacle-free trajectories if possible, and comes to a stop otherwise.

\subsection{Social Action Selection Benchmark Scenario}
\seclabel{benchmark}
As part of \approach{}, we include a complete benchmark for social navigation
action selection in a multipolicy setting similar to the one employed for
~\cite{holtz2021idips, mpdm2}. At each timestep, the agent is responsible for
choosing one action
from a discrete set of actions representing the existing sub-policies of a
navigation behavior. These sub-policies are designed as robust behaviors
built on the deterministic navigation module presented in \secref{navigation}.

A given benchmark in \approach{} is defined by the choice of
module implementations, action space,
observation space, and reward function and can be additionally customized
for complexity and difficulty via the configuration parameters of the underlying
modules. Examples of these configurable parameters include
\begin{paraenum}
  \item the environment map as defined by the environment module,
  \item the parameters of the humans,
  \item the robot motion model and kinematics,
  \item the amount of observation or execution noise in the simulation.
\end{paraenum}
In the following, we will describe the observation, action, and reward spaces
for the Social Action Selection Benchmark. Other benchmarks could be designed by
replacing one or more of the components described here.
For more information on the underlying
configuration of the submodules, we refer the reader to the supplementary material
and implementation.

We define the discrete action space to contain four sub-policies:
stopping in place (Halt), navigating towards
the next goal (GoAlone), following a human (Follow), and
passing a human (Pass). The observation space $\langle p_g, p_l, v_r, H_r \rangle$
consists of the global and local goal poses $p_g$ and $p_l$, respectively, the
robot velocity $v_r$, and the relative poses and velocities of the humans
visible to the robot $H_r$, with all non-visible human poses and velocities
set to 0. As such, the observation function needs to zero out the observations
of all humans that are \textbf{not} visible to the robot because they are
occluded by static obstacles in the environment. We choose this observation space
to minimize the environment-specific features in the observation by
using only local coordinates and velocities to minimize the risk of overfitting
to specific training examples.

Finally, the reward function is a weighted linear combination
of three metrics from Multipolicy Decision Making \cite{mpdm2}: $d_g$ the distance gained towards the goal since the last
timestep, $f$ the maximum force between the robot and a human, and finally,
$b$ the maximum blame resulting from the robot's actions and the humans in
the environment. Additionally, the agent receives a bonus $c$ if the goal is
reached during a timestep.
This reward function can be represented using weights $w_i$ as:
$r = (w_1 * d_g + w_2 * f + w_3 * b) + \texttt{if (success) c else } 0.0 $.
The weights can be used to specify user or task-specific preferences with
respect to the tradeoff between social compliance and time to goal.


\seclabel{technical}

\section{Benchmark and evaluation}
\seclabel{evaluation}
To demonstrate results obtained with \approach{} we provide experimental
results using the benchmark scenario described in \secref{benchmark},
and a series of baseline learned and engineered policies.
Our reference implementations include two
Reinforcement Learning (RL) implementations from the
Stable-Baselines project~\cite{stable-baselines3}, Proximal Policy Optimization (PPO)
and Deep Q-Networks (DQN),
two Learning from Demonstration (LfD) approaches, Behavior Cloning (BC) and
Generative Adversarial Imitation Learning (GAIL)
from the HumanCompatibleAI project~\cite{wang2020imitation},
as well as two engineered navigation
solutions, trajectory rollout based navigation and a social reference
solution (Ref)
and a symbolic learning from demonstration approach called Physics Informed Program
Synthesis (PIPS)~\cite{holtz2021idips}.
We evaluate the learned techniques on three criteria in three sets of experiments.
First we evaluate the training efficiency of each technique by comparing
the number of steps and amount of data needed for learning reasonable policies
in \secref{dataEfficiency}. Then we evaluate the performance of each policy
in terms of the social metrics defined in \secref{gym}, and compare the
learned policies to the engineered policies in \secref{performance}. Finally,
we consider the generalizability of the learned policies by transferring them
to a new simulated environment and comparing their performance in \secref{transfer}.

\begin{figure*}[ht]
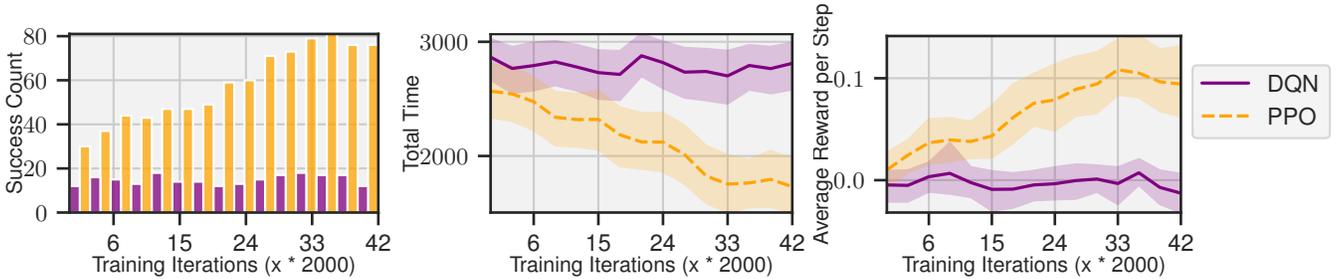

  \centering
  \hspace{-0.1\textwidth}
  \input{images/phase2/rl_success.pgf}
  \input{images/phase2/rl_time.pgf}
  \begin{subfigure}[t]{0.3\textwidth}
  \input{images/phase2/rl_reward.pgf}
  \end{subfigure}
  \vspace{-0.5em}
    \caption{Progressive performance with more training iterations.
    The shaded regions represent
    the 90\% confidence interval.}
  \figlabel{rlTraining}
  \vspace{-2em}
\end{figure*}

\subsection{Data Efficiency and Learning Rates}
\seclabel{dataEfficiency}
The learning algorithms we evaluate fit into two major categories that
each require different learning procedures.
Reinforcement learning approaches learn via interactive interaction
with the environment and Learning from Demonstration approaches require
apriori demonstrations of the desired behavior. For each set of
techniques, we evaluate and compare
the training process, learning rate, and data efficiency during learning
procedures utilizing \approach{} in the following paragraphs.

\paragraph{Reinforcement Learning}
Training new RL algorithms is a plug-and-play process that enables
the agent to interact with \approach{} and iteratively update its policy.
To evaluate the sample efficiency of the considered RL approaches, we trained
each model for $84000$ total timesteps and created a checkpoint of the learning
progress every $2000$ steps. For each model, we
evaluate their performance on the same $100$ test scenarios and
report their performance in terms of the reward described in \secref{benchmark}
in \figref{rlTraining}.
Our evaluation results show two key results, first, both RL algorithms
require $84000$ or more timesteps before peaking in performance, and second
the PPO approach is significantly more data efficient than the DQN approach
we evaluated. In general, the DQN approach is particularly weak to the
initial conditions, and performs more poorly than the other evaluated
approaches on the small sample of $100$ trials used for these experiments.

\paragraph{Learning from Demonstration}
To evaluate the data efficiency of
the considered
LfD approaches, we generated a single demonstration set with $155483$ timesteps from
our reference behavior
and trained multiple policies with decreasing quantities of
data evaluating each model on $100$ test scenarios, as shown in \figref{lfdTable}.
We report the performance in terms of the success rate and reward earned.
Notably, PIPS cannot learn policies with more than $3500$ data points,
as the symbolic synthesis approach used by PIPS does not scale to large numbers
of demonstrations. Conversely, while the DNN-based approaches can be trained
with $3500$ examples or less, their performance decreases significantly as the amount
of data used for training is reduced. Only performing similarly to the PIPS-based
policies when $155400$ samples are used.
There are two likely reasons for this
improved data-efficiency.
First, the
structure used for synthesis provides additional constraints,
and second, PIPS selectively subsamples the data
by windowing around transitions between actions, an optimization that is
not part of the more general purpose neural approaches.

\begin{figure}[htb]
  \begin{center}
  \footnotesize
\renewcommand{\arraystretch}{0.5}
    \begin{tabular}{rcccccc }
      \toprule
      \multirow{2}{*}{\textbf{Training Size}} &
      \multicolumn{2}{c}{\textbf{BC}} &
      \multicolumn{2}{c}{\textbf{GAIL}} &
      \multicolumn{2}{c}{\textbf{PIPS}}  \\
      \cmidrule(lr){2-7}
                 & \textbf{($\%$)} & \textbf{r}
                 & \textbf{($\%$)} & \textbf{r}
                 & \textbf{($\%$)} & \textbf{r} \\
      \midrule
      155400 & 84 & 0.12 & 59 & 0.06 & - & - \\
      116600 & 16 & 0.01 & 40 & 0.04 & - & - \\
      77700 & 37 & 0.04 & 38 & 0.02 & - & - \\
      38800 & 70 & 0.08 & 21 & 0.02 & - & - \\
      3500 & 49 & 0.05 & 27 & 0.01 & 88 & 0.16 \\
      \bottomrule
    \end{tabular}
  \vspace{-.5em}
  \caption{Performance of LfD approaches with varying numbers of demonstrations
    in terms of the success rate \textbf{($\%$)} and the average timestep
    reward $\textbf{r}$.}
  \figlabel{lfdTable}
  \end{center}
  \vspace{-2.0em}
\end{figure}


\begin{figure*}[!htb]
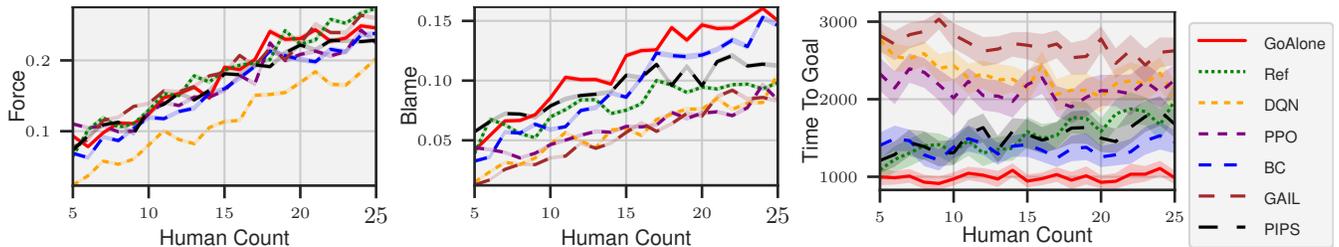

  \centering
  \hspace{-0.1\textwidth}
  \input{images/phase1/force.pgf}
  \input{images/phase1/blame.pgf}
  \begin{subfigure}[t]{0.3\textwidth}
  \input{images/phase1/time.pgf}
  \end{subfigure}
  \vspace{-0.5em}
    \caption{Performance of evaluated policies in the training environment.
    The shaded regions represent the 90\% confidence interval.}
  \figlabel{evalResults}
  \vspace{-1em}
\end{figure*}

\begin{figure*}[!htb]
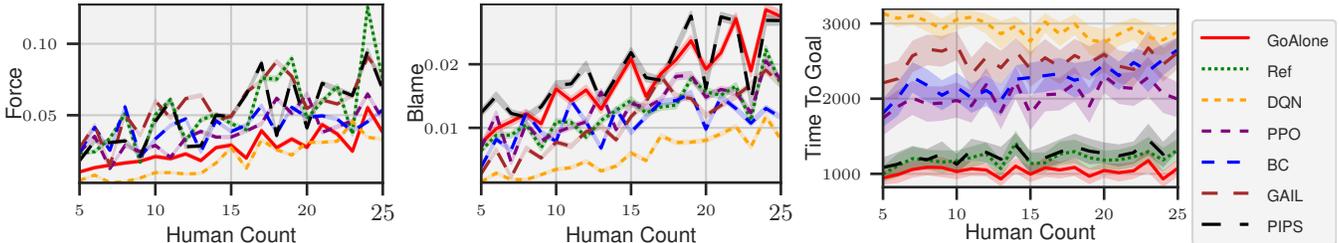

  \centering
  \hspace{-0.1\textwidth}
  \input{images/phase3/force.pgf}
  \input{images/phase3/blame.pgf}
  \begin{subfigure}[t]{0.3\textwidth}
  \input{images/phase3/time.pgf}
  \end{subfigure}
  \vspace{-0.5em}
    \caption{Performance of evaluated policies in a novel environment. The shaded regions represent
    the 90\% confidence interval.}
  \figlabel{transferResults}
  \vspace{-1em}
\end{figure*}

\begin{figure}[htb]
  \vspace{-1.0em}
  \begin{center}
  \footnotesize
\renewcommand{\arraystretch}{0.5}
    \begin{tabular}{lcc }
      \toprule
      \multirow{2}{*}{\textbf{Policy}} &
      \multicolumn{2}{c}{\textbf{Success Rates ($\%$)}}  \\
      \cmidrule(lr){2-3}
                 & \textbf{Test} & \textbf{Transfer}\\
      \midrule
      GoAlone & 97.5 & 97.3 \\
      \midrule
      Ref     & 89.4 & 92.4  \\
      \midrule
      BC   & 92.3 & 49.9  \\
      \midrule
      GAIL & 83.3 & 18.9 \\
      \midrule
      PIPS & 88.9 & 89.2 \\
      \midrule
      DQN & 54.1 & 8.8 \\
      \midrule
      PPO & 55.1 & 47.2 \\
      \bottomrule
    \end{tabular}
  \vspace{-.5em}
  \caption{Success rates in the training environment and after transferring to
    a novel environment.}
  \figlabel{successTable}
  \end{center}
  \vspace{-2.0em}
\end{figure}


\begin{figure}[ht]
  \centering
    \begin{subfigure}{0.49\columnwidth}
        \centering
        \includegraphics[width=0.99\textwidth]{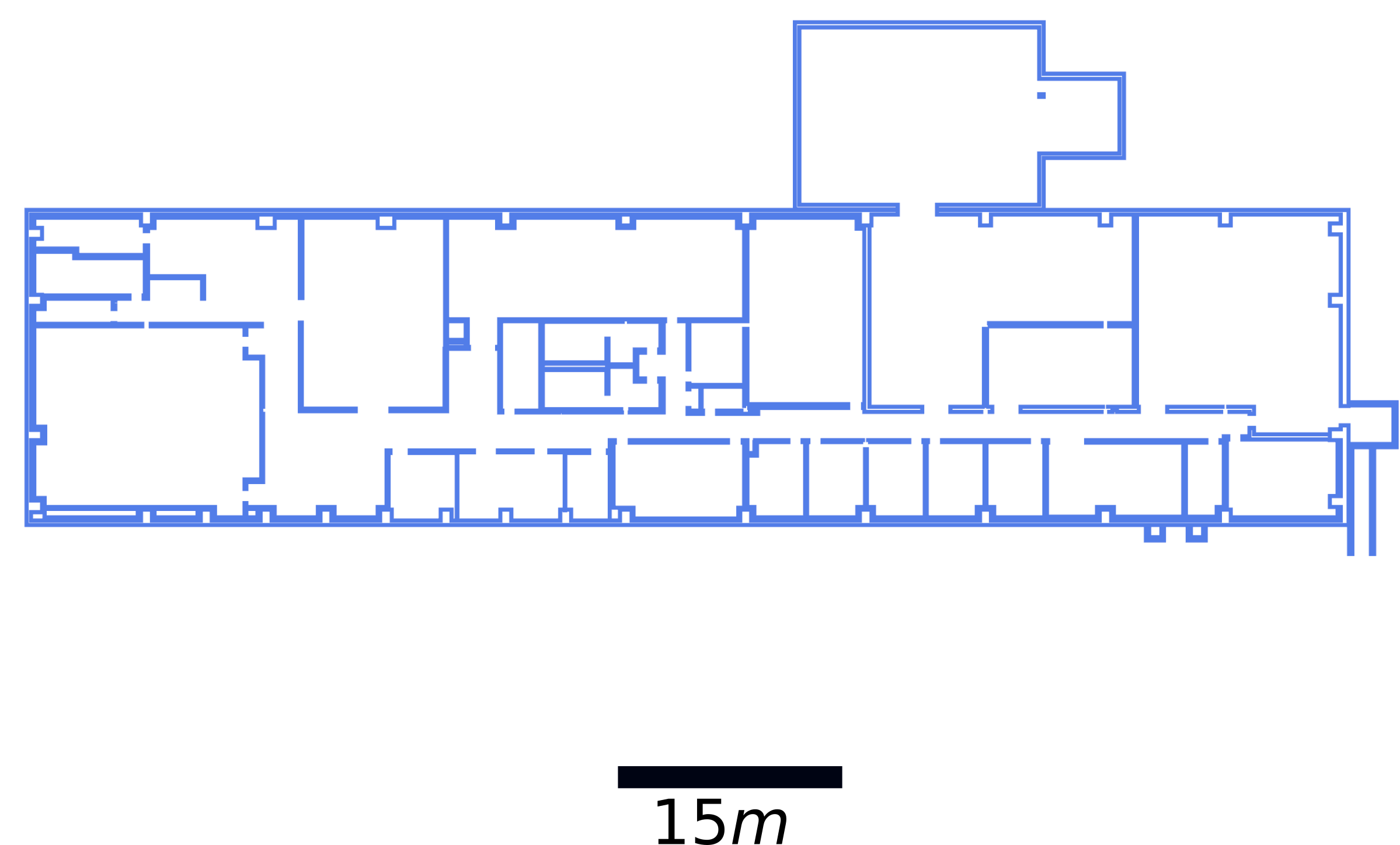}
    \end{subfigure}
    \begin{subfigure}{0.49\columnwidth}
        \centering
        \includegraphics[width=0.99\textwidth]{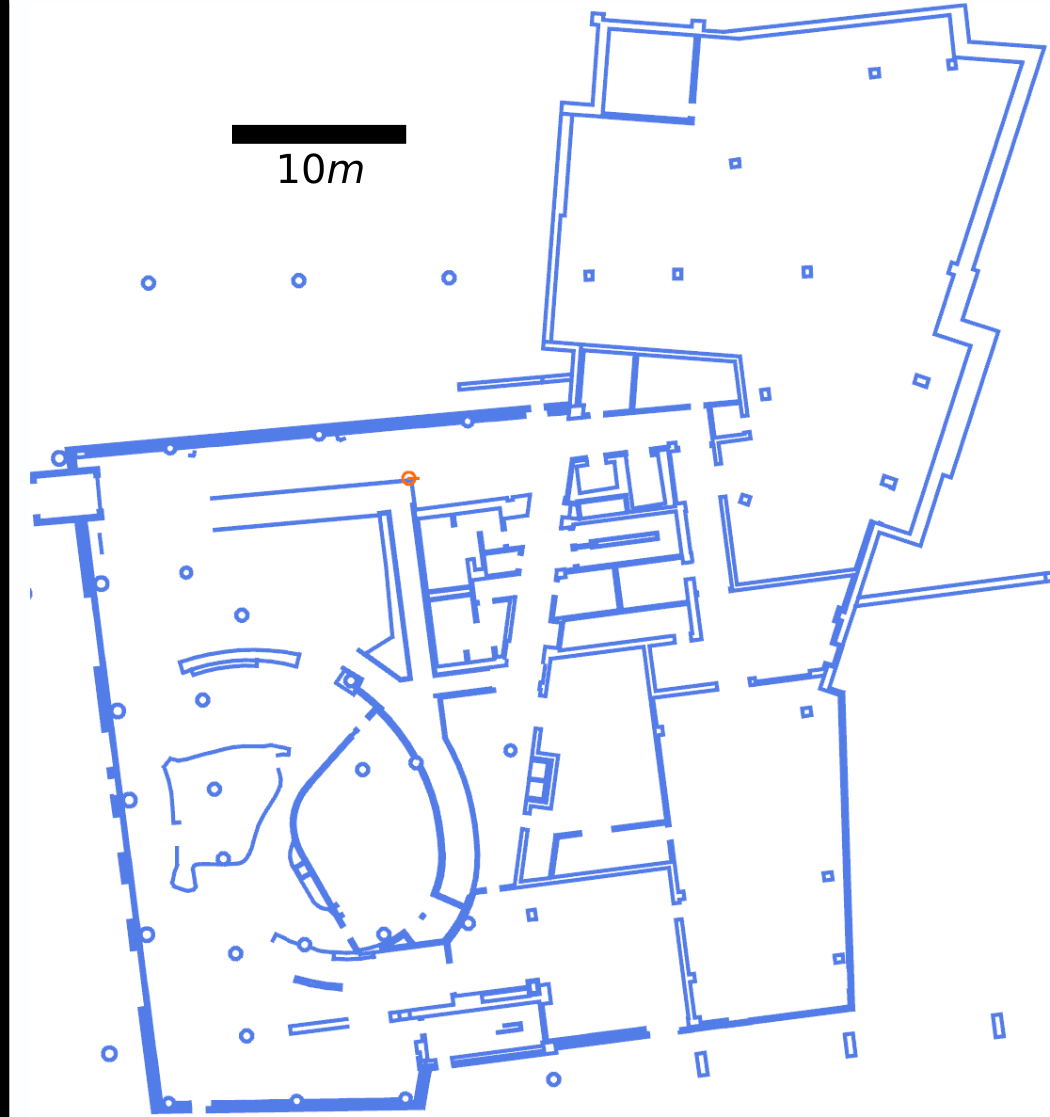}
    \end{subfigure}
    \caption{Environment used for training and evaluation on the left,
    and new environment for transfer evaluation on the right.}
  \figlabel{transferEnviron}
  \vspace{-1.0em}
\end{figure}

\subsection{Performance of Learned Models}
\seclabel{performance}
To evaluate and compare the learned behaviors, we evaluate them
on the same randomly generated set of $2000$ trials.
We report the results in terms of four primary metrics,
successful trials, force,
blame, and time to goal as described in \secref{gym}.
We report these metrics
as opposed to the score or reward to better compare between
LfD and RL approaches, and to better position the results with
respect to task performance. We report the percentage of successful trials
as a table in \figref{successTable} and the metric results in \figref{evalResults}.

In addition to the learned behaviors, \figref{successTable} and \figref{evalResults}
feature the engineered GoAlone behavior
utilizing only trajectory rollout as described in \secref{navigation}, and
the engineered Reference behavior (Ref) used for comparison and
generating demonstrations for the LfD algorithms. In \figref{successTable}, a trial
is counted as successful if the goal is reached within a bounded time without
any collisions between the robot and humans or walls in the environment.
In terms of pure success rate, no social behavior is as efficient as the
GoAlone policy, likely due to the halting robot problem, where the robot
is often left waiting for humans to pass long enough that the scenario times
out. When comparing the learned behaviors, we find that the LfD behaviors
have a better overall success rate than the RL approaches, suggesting that continuous
scenario demonstrations better convey the delayed reward of reaching the goal than
the reward function. When we look at the metric performance of each approach,
we see the success rates reflected in the time to goal, with the more successful
approaches featuring an overall lower time to goal, while the less successful
approaches are slower. An exception to this is the GAIL policy, which
was neither the most or least successful while also being the slowest policy.
As would be expected, we see a tradeoff
between time to goal and social compliance reflected to some degree in both the
force and blame graphs, with the slower behaviors exhibiting lower force
and blame. In particular, BC and PIPS roughly match the performance of the
Ref behavior from which demonstrations were drawn in terms of all three metrics,
while in contrast, GAIL, PPO, and DQN optimized for social compliance
over time to goal.

\subsection{Environment Transfer Performance}
\seclabel{transfer}
To evaluate the ability of the learned policies to transfer between environments,
we introduce a new set of $2000$ scenarios using a novel map with a significantly
different configuration than the map used for policy training. This new map
consists of a different set of static obstacles and a new set of possible
start and goal locations for both the robot and the humans.
For comparison, the environment used for the training and the evaluation
environment used for this experiment are shown in \figref{transferEnviron}.

We evaluate each policy on the same $2000$ random trials drawn from the new
environment and report the
performance in terms of force, blame, and time to goal in \figref{transferResults}
and in terms of the success rate in \figref{successTable}. The first key result
is that the engineered policies (Goalone, Ref) do not degrade in success rate during
the transfer between environments, while all but the PIPS-learned policy degrade
significantly.
Similarly, the time to goal metric reflects this, with more successful behaviors
achieving lower average times to goal and behaviors more likely to fail achieving
much slower average times to goal.
The new environment is less difficult for the engineered
policies, as the distribution of humans is sparser in the larger, more open, map.
This suggests that despite our efforts to design an observation space
and reward that are environment-agnostic, the non-symbolic learning algorithms
are still making decisions that are highly environment-dependent.
In terms of the social metrics, the results for the learned policies reflect
what we stated about the engineered policies. The new environment is more manageable
 with lower force and blame achieved by all behaviors thanks to the
more sparse human distributions. These results demonstrate what we would expect,
that the model-based
and symbolic approaches perform better in environment transfers than black-box
model-free approaches.

\section{Discussion and future work}
\seclabel{conclusion}
In this paper, we presented \approach{}, a configurable simulator and benchmarking
tool for socially navigating robots. \approach{} provides an interface for
easily integrating learning algorithms with a 2D simulator designed to abstract
away perception and localization and focus on the difficult task of learning
robust robot behaviors for navigating amongst humans in complex environments.
Further, we present empirical results from evaluating a suite of baseline algorithms in
our provided social action selection benchmark that demonstrate the utility
of \approach{}. While \approach{} provides an extensible interface, it is
difficult to imagine that any one solution will perfectly fit the needs of
myriad researchers, and we provide only a single initial benchmark evaluation
case initially. It is our hope that \approach{} will reduce the barrier to entry
for evaluating learning algorithms for social robot navigation by allowing
other researchers to build on an extensible foundation for social navigation
learning and evaluation.

\section{Acknowledgments}
This work is supported in part by NSF (CAREER2046955, SHF-2006404), ARO (W911NF-19-20333, W911NF-21-20217), and Northrop Grumman Mission Systems.

\bibliographystyle{IEEEtran}
\bibliography{references}

\end{document}